\documentclass[runninghead]{llncs}
\usepackage[T1]{fontenc}
\usepackage{xcolor}
\usepackage{hyperref}
\usepackage{graphicx}
\usepackage{amsmath}
\usepackage{listings}
\usepackage{booktabs}
\usepackage{algorithm}
\usepackage{algpseudocode}
\usepackage{array}

\hypersetup{
    colorlinks=true,
    linkcolor=blue,
    urlcolor=cyan
}
\begin{document}
\title{LMRPA: Large Language Model-Driven Efficient Robotic Process Automation for OCR}
\author{
    Osama Hosam Abdellatif \and
    Abdelrahman Nader Hassan \and
    Ali Hamdi
}
\institute{
    Faculty of Computer Science \\
    MSA University, Cairo, Egypt \\
    \email{\{osama.hosam, abdelrahman.nader, ahamdi\}@msa.edu.eg}
}
\maketitle

\begin{abstract}

This paper introduces LMRPA, a novel Large Model-Driven Robotic Process Automation (RPA) model designed to  greatly improve the efficiency and speed of Optical Character Recognition (OCR) tasks. Traditional RPA platforms often suffer from performance bottlenecks when handling high-volume repetitive processes like OCR, leading to a less efficient and more time-consuming process. LMRPA allows the integration of Large Language Models (LLMs) to improve the accuracy and readability of extracted text, overcoming the challenges posed by ambiguous characters and complex text structures.Extensive benchmarks were conducted comparing LMRPA to leading RPA platforms, including UiPath and Automation Anywhere, using OCR engines like Tesseract and DocTR. The results are that LMRPA achieves superior performance, cutting the processing times by up to 52\%. For instance, in Batch 2 of the Tesseract OCR task, LMRPA completed the process in 9.8 seconds, where UiPath finished in 18.1 seconds and Automation Anywhere finished in 18.7 seconds. Similar improvements were observed with DocTR, where LMRPA outperformed other automation tools conducting the same process by completing tasks in 12.7 seconds, while competitors took over 20 seconds to do the same.  These findings highlight the potential of LMRPA to revolutionize OCR-driven automation processes, offering a more efficient and effective alternative solution to the existing state-of-the-art RPA models.

\end{abstract}

\section{INTRODUCTION}
In today’s rapidly evolving digital landscape, businesses seek to replace mundane and time-consuming tasks processes with automation to make operations very efficient and cost-effective. Robotic Process Automation (RPA) is an emerging technology of these days, which is being used by organizations for streamlining workflows by replicating human interactions with digital systems using software-based robots \cite{madakam2019future}. These robots can perform a wide variety of tasks, such as sending emails, extracting data from web pages, managing files, and logging into applications, with precision and consistency. The motivation behind RPA lies in its ability to free up human resources from mundane, rule-based tasks, enabling them to focus on higher-value activities while reducing the risk of human error. This makes RPA a key enabler of digital transformation in modern businesses, significantly improving productivity and efficiency \cite{tripathi2018learning}.

Traditional RPA systems are mainly rule-driven and perform well in automating simple, repetitive tasks. However, they usually struggle with more complex processes that involve unstructured data, such as text recognition from documents or images. Optical Character Recognition (OCR) \cite{al2018arabic,hamdi2021c,al2018enhanced} has now become an integral part of most automation workflows, yet the integration of OCR with RPA faces challenges regarding accuracy, speed, and adaptability \cite{baweja2023comparative}. The currently available RPA platforms ,such as UiPath and Automation Anywhere, works well for general automation tasks, but usually present some bottleneck performances when dealing with high-volume, repetitive OCR tasks, leading to inefficiencies and delays \cite{uipath2024platform} \cite{automation_anywhere2024advanced}. These limitations highlight the need for more advanced RPA models capable of managing complex, unstructured files while maintaining high efficiency and accuracy.

In this regard, this paper proposes LMRPA,a novel large model-driven robotic process automation model specifically designed to enhance the efficiency and accuracy of OCR tasks within automation workflows. LMRPA thus uses the power of Large Language Models (LLMs) to enhance text extraction performance by correctly interpreting ambiguous characters and complex text structures, resulting in more accurate and reliable OCR outputs. Our model introduces a novel architecture that integrates LLMs into the RPA process, allowing for processing of OCR data towards error minimization and enhancing performance. Through extensive benchmarking, we demonstrate that LMRPA outperforms existing RPA platforms \cite{moffitt2018robotic} \cite{mullakara2024selecting}, such as UiPath and Automation Anywhere, in both speed and efficiency, particularly when dealing with large-scale OCR tasks. This research showcases the potential of LMRPA to revolutionize automation processes, making them faster, more accurate, and more adaptable to complex scenarios.

\subsection{Research Problem and Contributions}
This paper investigates the following problem: a shortage of efficiency and performance bottlenecks when Optical Character Recognition tasks are handled using existing Robotic Process Automation systems. Even with fast development and expansion in recent times, existing RPA systems\cite{agarwal2023rpa}
 struggle to manage unstructured data and complex structures of texts, resulting in delays and errors and therefore reduced operational efficiency.

This paper proposes LMRPA, a novel approach leveraging Large Language Models to optimize OCR processes within RPA systems. The key contributions of this work include: 
\begin{itemize}
    \item \textbf{Integration of Large Language Models with RPA for OCR}: Introducing a new architecture that applies LLMs for pre-processing and error reduction in OCR workflows.
    \item \textbf{Performance Benchmarking}: Conducting extensive benchmarking experiments to demonstrate that LMRPA outperforms existing RPA platforms, including UiPath and Automation Anywhere, in terms of speed and efficiency, particularly for high-volume OCR tasks.
    \item \textbf{LMRPA Model Implications}: Highlighting the potential of LMRPA to significantly improve OCR processes, making them faster, more accurate, and adaptable to various real-world applications.
\end{itemize}

The rest of this paper is organised as follows. Section \ref{rw} highlights the related work in RPA, OCR, and LLM. Section \ref{m} discusses the methodology and experiments. Section \ref{rd} interprets the results and discusses their implications. Section \ref{c} summarizes the key findings and contributions.

\section{Related Work}\label{rw}
In today’s digital age, businesses\cite{hussain2023rpa}
 are under increasing pressure to enhance operational efficiency and reduce costs. One of the most transformative technologies emerging to meet these challenges is Robotic Process Automation (RPA), which automates repetitive tasks that are typically time-consuming and error-prone. RPA uses software robots to replicate human actions in digital systems, such as extracting data, managing files, and interacting with applications. These robots can work around the clock without human intervention, freeing up human resources to focus on more complex, value-added activities. Tools like UiPath and Automation Anywhere are at the forefront of RPA adoption, particularly in automating processes that require Optical Character Recognition (OCR) — a technology that allows machines to read and interpret text from scanned documents and images.

While traditional RPA systems have proven to be highly effective at automating structured, repetitive tasks, they often struggle when faced with unstructured data, such as handwritten text or complex document layouts. OCR has become a critical component in bridging this gap, but it also introduces its own set of challenges. For instance, traditional OCR systems may struggle with ambiguous characters or non-standard fonts, leading to errors in text recognition. To address these challenges, businesses have increasingly turned to Large Language Models (LLMs), a class of artificial intelligence models\cite{cfbbots2018rpaai} capable of interpreting and understanding language in ways that go beyond simple pattern matching. By incorporating LLMs into OCR workflows, it’s possible to enhance accuracy, especially in dealing with unstructured or complex data.

Integrating LLMs into RPA tools offers significant advantages. These models can be trained to recognize ambiguous characters, disambiguate complex sentence structures, and adapt to various languages and fonts. For example, Google OCR and Microsoft OCR are integrated into RPA tools like UiPath, enabling them to read and process documents more accurately. However, the efficiency of these tools is often limited by the underlying OCR engines, which still struggle to provide accurate results in the presence of noise or distorted text. LLMs, with their ability to contextualize text, offer a powerful solution to these limitations by improving OCR output quality. This synergy between LLMs and OCR is an area that remains underexplored, and it’s where new advancements could unlock even greater potential for automation.

Despite the progress made in the field, several gaps remain in the research. Many studies on RPA and OCR lack transparent methodologies, making replication difficult. This is particularly evident in studies that do not disclose the datasets used or the specific configurations of OCR engines tested. Such omissions hinder the reproducibility of results and limit the ability to compare the effectiveness of different RPA tools\cite{baweja2023rpatools}. Moreover, while much has been written about the advantages of using RPA and LLMs in isolation, few studies have compared the performance of these tools against each other in real-world, large-scale OCR tasks. Furthermore, critical metrics such as processing speed, scalability, and error rates are often overlooked, despite their importance for practical deployment in business environments.

Our research aims to fill these gaps by investigating how LLMs can be integrated into RPA workflows to improve OCR tasks. By conducting thorough benchmarking, we seek to provide a clearer picture of the performance improvements that can be achieved through this integration. In particular, we aim to quantify the efficiency gains, identify any potential trade-offs, and explore how this enhanced system compares to current industry\cite{vialle2020rpa} leaders like UiPath and Automation Anywhere. Ultimately, our goal is to offer a more efficient and accurate solution for OCR-driven automation, one that addresses the challenges of unstructured data while providing the transparency and reproducibility that current studies often lack. Through this work, we hope to lay the groundwork for future innovations in RPA and LLM integration, advancing the field and providing businesses with a more powerful tool for digital transformation.


\section{Research Methodology}\label{m}
We design the LMRPA model to continuously monitor a designated directory for new invoice files, process detected images using OCR engines \cite{sharma2022applications}, and refine the extracted data using a Large Language Model (LLM)\cite{smith2024gemini} to generate structured JSON \cite{ferreira2024evaluation}.Further, The system then populates the database with the structured data and automatically generates a formatted Template Report, operating in a fully automated loop \cite{moffitt2018robotic}.

\begin{figure}[ht]
    \includegraphics[width=0.9\textwidth]{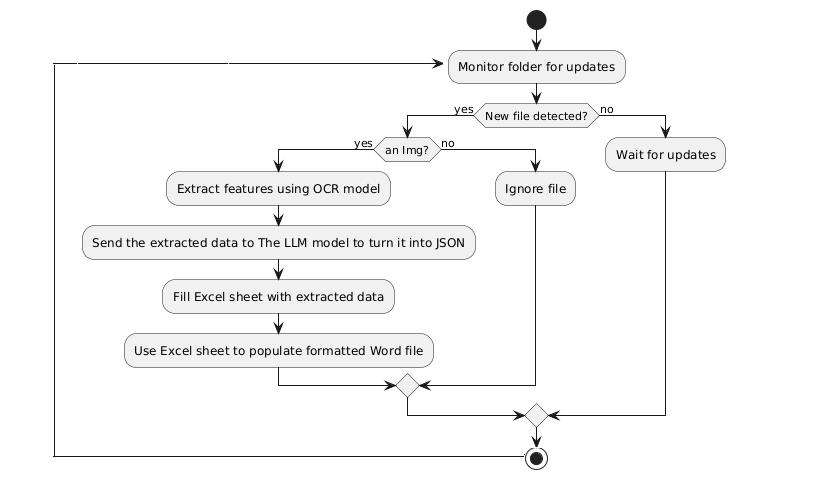} 
    \caption{The proposed LMRPA system architecture.}
    \label{fig1}
\end{figure}

\subsection{LMRPA Model}

\subsubsection{Process Monitoring Function}
The proposed \textit{LMRPA} system (as shown in \textbf{Figure 1}) begins by continuously monitoring a designated folder $\mathcal{D}$ for new files. Let $\mathcal{F}(t)$ represent the set of files at time $t$. The system detects a new file $f_i$ if:
\[
f_i \in \mathcal{F}(t) \setminus \mathcal{F}(t - \Delta t)
\]
where $\Delta t$ is the monitoring interval. If $f_i$ is detected and belongs to a valid set of image files $\mathcal{I}$ (i.e., $f_i \in \mathcal{I}$), the file proceeds for processing; otherwise, it is ignored.

\subsubsection{OCR Extraction}
When a valid image file $f_i \in \mathcal{I}$ is detected, the system applies an OCR engine $\mathcal{O}_k$ to extract textual data $\mathbf{T}$ from the image. The OCR engine can be either Tesseract \cite{tesseract2024ocr}($\mathcal{O}_1$) or docTR \cite{mullakara2024selecting}($\mathcal{O}_2$). The extracted text $\mathbf{T}$ is given by:
\[
\mathbf{T} = \mathcal{O}_k(f_i), \quad k \in \{1, 2\}
\]

\subsubsection{LLM for Data Structuring}
The raw text $\mathbf{T}$ extracted by the OCR engine is sent to a Large Language Model (LLM) $\mathcal{L}$, which converts it into structured JSON data $\mathbf{J}$. This process is represented as:
\[
\mathbf{J} = \mathcal{L}(\mathbf{T})
\]
The structured JSON ensures the data conforms to the schema required by the system’s database and reporting tools\cite{johnson2023llmdataset,smith2024systemllm}
.

\subsubsection{Data Population and Report Generation}
The structured JSON data $\mathbf{J}$ is mapped into an Excel sheet $\mathcal{E}$ using a mapping function $\mathcal{M}$:
\[
\mathcal{E} = \mathcal{M}(\mathbf{J})
\]
The Excel sheet data is then used to generate a formatted Word document $\mathcal{W}$:
\[
\mathcal{W} = \mathcal{F}_{\text{doc}}(\mathcal{E})
\]

\subsubsection{Continuous Automation}
The system operates in a continuous loop, represented by the algorithm (see \textbf{Algorithm 1}), where it monitors for new files and processes them as they arrive:
\[
\text{while } t \in \mathcal{T}, \quad \text{perform LMRPA process}
\]

\subsection{The LMRPA Algorithm}
\begin{algorithm}[h]
\caption{The LMRPA Algorithm}
\begin{algorithmic}[1]
    \State \textbf{Initialize} monitoring of directory $\mathcal{D}$
    \While{True}
        \State \textbf{Check} for new files in $\mathcal{D}$
        \If{new file $f_i$ is detected}
            \If{$f_i \in \mathcal{I}$} \Comment{Check if $f_i$ is an image}
                \State Extract text data $\mathbf{T}$ using OCR engine $\mathcal{O}_k(f_i)$
                \State Send $\mathbf{T}$ to LLM $\mathcal{L}$ to create structured JSON $\mathbf{J}$
                \State Save $\mathbf{J}$ to the database $\mathit{DB}$
                \State Populate Excel sheet $\mathcal{E}$ and generate Word report $\mathcal{W}$
            \Else
                \State Ignore non-image file
            \EndIf
        \EndIf
    \EndWhile
\end{algorithmic}
\end{algorithm}

\section{Benchmark and Experimental Design}
The data used in this research was gathered from various sources, with a primary focus on invoice datasets. The utilised dataset, comprising 3,000 invoice images, was sourced from platforms like Kaggle and Roboflow \cite{roboflow2024invoice} and Kozłowski’s "Samples of Electronic Invoices" \cite{kozlowski2021samples}. The images were selected to ensure high-quality input for OCR processing. These datasets were selected to test the performance of the RPA tools, especially regarding image clarity and its effect on the efficiency of OCR.

The research employed a simulated invoice processing environment. All RPA tools were tested on the same dataset and device to ensure consistent comparisons. Two different OCR engines, Tesseract and docTR, were used to test the results of data extraction that each automation tool produces. This experimental setup aimed to provide measurable outcomes, focusing on metrics like speed . The experimental results were benchmarked against state-of-the-art (SOTA) practices in RPA, focusing directly on aspects such as operational efficiency and performance.

Potential limitations in the research methodology including the exclusion of Blue Prism from the comparative analysis due to the lack of a community version, which may affect the comprehensiveness of the findings. Moreover,using community versions may not be representative or reflective of the capabilities of commercial tools. Further, there was the introduction of a 5-second delay to adhere to limited API access, and the dataset was split into batches of 1,500 images to adhere to daily usage limits.

\section{Results and Discussion}\label{rd}

Table \ref{tab} summarizes the performance comparison of our custom LMRPA model against the state-of-the-art RPA tools, UiPath and Automation Anywhere, across different batches using Tesseract and DocTR as OCR engines.

\begin{table}[h]
\centering
\caption{Comparison of Automation Models by Task}
\label{tab}
\begin{tabular}{lcc|cc}
\toprule
\textbf{Task} & \multicolumn{2}{c}{\textbf{Tesseract}} & \multicolumn{2}{c}{\textbf{DocTR}} \\ 
\cmidrule(lr){2-3} \cmidrule(lr){4-5}
& \textbf{Batch 1} & \textbf{Batch 2} & \textbf{Batch 1} & \textbf{Batch 2} \\ 
\midrule
\textbf{UiPath} & 18.1 sec & 18.0 sec & 21.4 sec & 20.1 sec \\ 
\textbf{Automation Anywhere} & 18.7 sec & 18.3 sec & 22.0 sec & 20.6 sec \\ 
\textbf{LMRPA (Ours)} & 9.8 sec & 9.4 sec & 12.7 sec & 12.4 sec \\ 
\bottomrule
\end{tabular}
\end{table}

\subsection*{Performance Comparison (Tesseract OCR)}

For Tesseract OCR in \textbf{Batch 1}, which involved larger image sizes (such as invoices), LMRPA was the fastest, processing the task in 9.8 seconds. In comparison, UiPath and Automation Anywhere were considerably slower, with processing times of 18.1 seconds and 18.7 seconds, respectively. These results highlight the efficiency of our custom solution, particularly when dealing with large image sizes.

In \textbf{Batch 2}, which involved smaller images, all tools showed slight improvements in processing times. LMRPA continued to lead with the fastest time of 9.4 seconds, followed by UiPath at 18.0 seconds and Automation Anywhere at 18.3 seconds. While the smaller image sizes contributed to faster processing times, LMRPA maintained a clear performance lead.

\subsection*{Performance Comparison (DocTR OCR)}

Switching to \textbf{DocTR OCR} in \textbf{Batch 1} resulted in an increase in processing times across all tools, due to the more computationally intensive nature of the DocTR engine. Nevertheless, LMRPA still outperformed both UiPath and Automation Anywhere, completing the task in 12.7 seconds. UiPath took 21.4 seconds, and Automation Anywhere took 22.0 seconds. While DocTR is slower than Tesseract, it provides cleaner, more accurate outputs, making it more suitable for applications where accuracy is prioritized over speed.

In \textbf{Batch 2} with DocTR, processing times for all tools improved slightly. LMRPA finished processing in 12.4 seconds, while UiPath and Automation Anywhere took 20.1 seconds and 20.6 seconds, respectively. Although DocTR remains computationally intensive, LMRPA's performance excelled in both accuracy and speed, reinforcing its advantage over the other tools.

\subsection*{Key Findings}

From the results in Table \ref{tab}, it is evident that LMRPA consistently outperforms both UiPath and Automation Anywhere across different OCR engines . The custom LMRPA solution demonstrates a significant advantage in terms of speed and efficiency, particularly for high-volume, repetitive tasks such as invoice processing.

The study shows substantial productivity gains with LMRPA. For comparison, manual invoice processing typically takes up to \textbf{600 seconds}. With \textbf{Tesseract OCR}, LMRPA reduced this time to \textbf{9.8 seconds} in \textbf{Batch 1}, representing a \textbf{98.4\% reduction} in processing time compared to manual methods. In \textbf{Batch 2}, LMRPA processed invoices in \textbf{9.4 seconds}, further emphasizing its efficiency.

When using \textbf{DocTR}, while the processing time was higher compared to Tesseract, LMRPA still significantly outperformed manual processing. In \textbf{Batch 1}, LMRPA completed the task in \textbf{12.7 seconds}, while UiPath took \textbf{21.4 seconds} and Automation Anywhere took \textbf{22.0 seconds}, resulting in \textbf{59.3\%} and \textbf{57.7\%} time savings over UiPath and Automation Anywhere, respectively. Compared to manual processing, LMRPA achieved a \textbf{97.8\%} time saving. In \textbf{Batch 2}, LMRPA processed the tasks in \textbf{12.4 seconds}, saving \textbf{61.7\%} and \textbf{60.2\%} compared to UiPath and Automation Anywhere, respectively, and \textbf{97.93\%} compared to manual processing.

\subsection*{Discussion}

The results underscore LMRPA's ability to significantly improve the performance of OCR-based RPA tasks\cite{pekkola2017rpa}
, offering Higher speed over existing RPA tools like UiPath and Automation Anywhere. These findings are particularly relevant for industries dealing with high volumes of repetitive OCR tasks, such as invoice processing, where automation can yield substantial productivity and efficiency gains.

However, it is important to note that developing tailor-made RPA solutions with platforms like UiPath and Automation Anywhere involved a steep learning curve. While these platforms feature user-friendly interfaces, they require advanced expertise for handling non-standard or complex scenarios. Furthermore, RPA bots are sensitive to process changes, requiring constant monitoring and performance reporting. Exception handling and managing unexpected cases often involve complex coding or manual intervention, reducing the scalability and flexibility of RPA solutions in dynamic environments.

In conclusion, LMRPA's exceptional performance demonstrates the potential of customized automation solutions to drive significant improvements in OCR-related workflows, providing an efficient alternative to traditional methods and competing RPA platforms.

\section{Conclusion}\label{c}
This paper has provided a comprehensive analysis of UiPath, Automation Anywhere LMRPA, focusing on their integration with OCR technologies. Each tool offers unique advantages, making them suitable for different business environments and automation needs. UiPath stands out for its user-friendliness, various packages, and strong community support, Automation Anywhere for its cognitive capabilities and flexibility. LMRPA offers a highly customizable and flexible solution, ideal for businesses with specific and complex automation requirements.The integration of OCR with these tools significantly enhances their capabilities, allowing them to automate processes involving complex document and image recognition. The results show great promise for such technology to streamline many exhausting and repetitive tasks.Although businesses should carefully consider their specific process requirements and the unique features of each tool when selecting an RPA solution to achieve optimal efficiency and effectiveness in their automation initiatives.

\section{Data and Code Availability}
The dataset used in this research is available at the following link: 
\href{https://www.kaggle.com/datasets/osamahosamabdellatif/high-quality-invoice-images-for-ocr}{Data Set}.

\noindent The code repository for this project can be found at: \href{https://github.com/MAD-SAM22/ERPA-OCR}{Repo}.

\section*{Acknowledgment}

We extend our heartfelt gratitude to \href{https://msa.edu.eg/msauniversity/}{Modern Science and Arts University (MSA University)} for their funding and continuous support throughout this research. Their generous contribution has been instrumental in the successful completion of our work. We would also like to express our deep thanks to \href{https://aitech.net.au/}{AiTech company} for their invaluable support , guidance throughout and funding this research. Their expertise and assistance have played a crucial role in the development and success of this project.

\bibliographystyle{splncs04}
\bibliography{main}


\end{document}